%% file: main.tex
\pdfoutput=1

\documentclass[11pt]{article}

\usepackage[final]{acl}

\usepackage{times}
\usepackage{latexsym}

\usepackage[T1]{fontenc}

\usepackage[utf8]{inputenc}

\usepackage{microtype}

\usepackage{inconsolata}

\usepackage{graphicx}
\usepackage{hyperref}

\usepackage{xcolor}
\usepackage{xspace}

\usepackage{enumitem}
\usepackage{amsmath}
\usepackage{booktabs}
\usepackage{hyperref}

\setlist{leftmargin=*,nosep}

\title{Rethinking Legal Judgement Prediction in a Realistic Scenario in the Era of Large Language Models}

 \author{
   Shubham Kumar Nigam\textsuperscript{1} \quad
   Aniket Deroy\textsuperscript{2} \quad
   Subhankar Maity\textsuperscript{2}  \quad
   Arnab Bhattacharya\textsuperscript{1} \\
   \textsuperscript{1}Indian Institute of Technology Kanpur (IIT-K) \\
   \textsuperscript{2}Indian Institute of Technology Kharagpur (IIT-Kgp)\\
   \texttt{\{sknigam, arnabb\}@cse.iitk.ac.in} \\
   \texttt{\{roydanik18, subhankar.ai\}@kgpian.iitkgp.ac.in}
}

\begin{document}
\maketitle

\input{abstract}

\input{intro}

\input{related_work}

\input{task_description}

\input{dataset}

\input{methodology}

\input{results_and_analysis}

\input{conclusion_and_future_work}

\input{limitations}

\input{ethics}

\bibliography{custom}

\input{appendix}

\end{document}

%% file: abstract.tex
\begin{abstract}
This study investigates judgment prediction in a realistic scenario within the context of Indian judgments, utilizing a range of transformer-based models, including InLegalBERT, BERT, and XLNet, alongside LLMs such as Llama-2 and GPT-3.5 Turbo. In this realistic scenario, we simulate how judgments are predicted at the point when a case is presented for a decision in court, using only the information available at that time, such as the facts of the case, statutes, precedents, and arguments. This approach mimics real-world conditions, where decisions must be made without the benefit of hindsight, unlike retrospective analyses often found in previous studies. For transformer models, we experiment with hierarchical transformers and the summarization of judgment facts to optimize input for these models. Our experiments with LLMs reveal that GPT-3.5 Turbo excels in realistic scenarios, demonstrating robust performance in judgment prediction. Furthermore, incorporating additional legal information, such as statutes and precedents, significantly improves the outcome of the prediction task. The LLMs also provide explanations for their predictions. To evaluate the quality of these predictions and explanations, we introduce two human evaluation metrics: \textit{Clarity} and \textit{Linking}. Our findings from both automatic and human evaluations indicate that, despite advancements in LLMs, they are yet to achieve expert-level performance in judgment prediction and explanation tasks.
\end{abstract}


%% file: intro.tex
\section{Introduction}
Predicting case outcomes based on judge-summarized narratives is an important task. Unlike previous studies~\cite{malik-etal-2021-ildc, nigam-etal-2024-legal} and \cite{vats2023llms}, we aim to simulate realistic scenarios where legal judgment prediction systems are used to predict and explain judgments as cases arrive on the bench for adjudication. Our approach focuses on the core factual components of the case—specifically, the events that led to the case being filed, which serve as the basis for judgment prediction. These facts are the foundation of legal arguments and provide the context needed for making judicial decisions. In contrast to previous works that have included the entire case text (including proceedings), our focus on facts mirrors real-world conditions, where judges rely primarily on the case facts when delivering judgments.

In addition to the facts of the case, we incorporate additional legal information such as statutes, precedents, and arguments. Statutes represent codified legal principles, while precedents provide case-specific rulings that help guide decision-making. Together, these legal frameworks offer a structured basis upon which judges rely when formulating their rulings. By extracting and integrating these elements into our models, we aim to enhance both the prediction and explanation tasks by grounding the analysis in actual legal texts and the governing principles that are applied in real cases.

We explore the efficacy of various transformer-based models investigate the impact of summarizing legal judgments~\cite{xx4,xx9,xx10,xx11} using techniques~\cite{xx1,xx2,xx16,xx12} such as BERTSum~\cite{liu2019fine}, CaseSummarizer~\cite{polsley2016casesummarizer}, LetSum~\cite{farzindar2004atefeh}, and SummaRuNNer~\cite{nallapati2017summarunner}. Our findings suggest that leveraging summarized information yields decent results in judgment prediction. 

To further enhance the quality of prediction, we introduce hierarchical transformer models that utilize the entirety of judgment facts, demonstrating superior performance compared to traditional summarization methods. Additionally, our examination of LLMs, including Llama-2 (13b \& 70b) \cite{touvron2023llama} and GPT-3.5 Turbo \cite{brown2020language}, highlights the exceptional performance of GPT-3.5 Turbo in the context of Indian legal judgment prediction. We find that augmenting our models with additional legal information, such as statutes, precedents, and arguments, significantly improves the quality of both tasks.

In addition to focusing on the accuracy of legal judgment prediction, it is equally important to assess the quality of the explanations provided by the models. For this reason, we introduce two novel human evaluation metrics: \textit{Clarity} and \textit{Linking}. \textit{Clarity} refers to how well the predictions and explanations are structured and whether they convey the reasoning in a clear and understandable manner. This is critical in the legal domain, where complex legal concepts must be communicated effectively. \textit{Linking}, on the other hand, evaluates the logical consistency between the explanation and the final judgment. It assesses whether the explanation effectively ties back to the outcome and supports the predicted decision. These metrics are vital because, while models may produce accurate predictions, their explanations often lack coherence or fail to justify the decision meaningfully. By incorporating these metrics, we aim to ensure that models provide not only accurate outcomes but also transparent and interpretable explanations that can be trusted by legal professionals.

The key contributions of this study are:
\begin{enumerate}
    \item We focus on evaluating the performance of several transformer-based models and hierarchical transformer models, specifically on factual data, to mirror real-world conditions in judgment prediction. This approach contrasts with previous works that utilized full case texts.
    \item We utilize LLMs to assess their capabilities in legal judgment prediction and explanation tasks.
    \item We define two human evaluation metrics, \textit{Clarity} and \textit{Linking}, to assess the quality of LLM-generated judgment predictions and explanations, providing a comprehensive assessment of the overall task performance.
\end{enumerate}
To ensure reproducibility, both the code and dataset have been made publicly available via our repository\footnote{\href{https://github.com/ShubhamKumarNigam/Realistic_LJP}{https://github.com/ShubhamKumarNigam/Realistic\_LJP}}. Additionally, for convenience, we have uploaded the data\footnote{\href{https://huggingface.co/collections/L-NLProc/realistic-ljp-models-670ccbb2b390830b3989f6bf}{huggingface.co/collections/L-NLProc/Realistic\_LJP-models}} and models\footnote{\href{https://huggingface.co/collections/L-NLProc/realistic-ljp-datasets-670ccbeab5aea07a37e86df8}{huggingface.co/collections/L-NLProc/Realistic\_LJP-datasets}} to Huggingface.

%% file: related_work.tex
\section{Related Work}
The field of Legal Judgment Prediction (LJP) has seen significant advancements, driven by the need to automate legal case outcome forecasting and alleviate the burden of overwhelming caseloads. Early works by \cite{aletras2016predicting}, \cite{chalkidis2019neural}, and \cite{feng2021recommending} laid the foundation for LJP, emphasizing the importance of explainability in AI predictions. Benchmark datasets such as CAIL2018 \cite{xiao2018cail2018}, ECHR-CASES \cite{chalkidis2019neural}, and others have spurred research in this area, inspiring models like TopJudge and MLCP-NLN. However, there remains a gap between machine and human performance.

In the Indian context, datasets like ILDC \cite{malik-etal-2021-ildc}, PredEx \cite{nigam-etal-2024-legal} and \cite{nigam2022nigam, malik-etal-2022-semantic, nigam2023legal} have highlighted the growing role of AI in legal judgments, with an emphasis on explainability. Research in LJP with LLMs, such as \cite{vats2023llms} and \cite{nigam-etal-2024-legal}, has experimented with models like GPT-3.5 Turbo and Llama-2 on Indian legal datasets. Other studies, such as \cite{masala2021jurbert} on Romanian legal texts and \cite{hwang2022multi} on Korean legal language, have demonstrated LJP's adaptability across legal systems.

Cross-jurisdictional work, including \cite{zhao-etal-2018-learning}, showcases LJP's applicability in different legal frameworks, with research expanding to multilingual considerations, as seen in \cite{niklaus2021swiss} and \cite{kapoor-etal-2022-hldc} for Hindi legal documents. Recent innovations, such as event extraction and multi-stage learning~\cite{feng-etal-2022-legal}, continue to push the boundaries of LJP research.



%% file: task_description.tex
\section{Task Definition}
This study focuses on Supreme Court of India (SCI) judgments, and the Court Judgment Prediction with Explanation task consists of two subtasks:

\textbf{Task A: Judgment Prediction}: This subtask is framed as a binary classification problem specific to SCI cases. Given a segment of the legal judgment as input, the goal is to predict whether the decision favors or is against the appellant. The prediction is represented by binary labels: \{1, 0\}, where 1 indicates that the appeal is accepted (i.e., if any part of the appeal is accepted, the decision is considered in favor of the appellant). Although some cases might involve multiple heads of appeal, where an appellant might win on some grounds and lose on others, for the purposes of this task, the outcome is simplified to a binary decision. Cases with mixed outcomes are excluded or reduced to this binary format for prediction.

\textbf{Task B: Rationale Explanation}: This subtask involves generating a coherent explanation or rationale that justifies the predicted decision, based on the provided segment of the judgment. The explanation seeks to clarify the reasoning behind the predicted outcome.
 
The workflow of the system, as illustrated in Figure~\ref{fig1}, captures the entire process—from extracting facts and additional legal information (such as statutes, precedents, and lower court rulings) to feeding this data into transformer models, hierarchical transformers, and LLMs. The diagram visually represents the pipeline of both tasks, highlighting how the prediction and explanation processes interact to form a comprehensive legal judgment prediction system.

%% file: dataset.tex
\section{Dataset}
We utilize the ILDC-multi dataset, as described by ~\cite{malik-etal-2021-ildc}, which comprises a total of 34,816 legal judgments from the Supreme Court of India, collected from 1947 to April 2020 via the Indian Kanoon website\footnote{\url{https://indiankanoon.org/}}. This dataset is divided into three subsets: training, validation, and test which contains 32,305, 994, and 1,517 judgments correspondingly. It is specifically designed to support the tasks of Court Judgment Prediction and Explanation (CJPE), with a portion of the legal judgment serving as input for both prediction and explanation processes. Additionally, a subset of this corpus is annotated with gold-standard explanations provided by legal experts, enhancing its utility for developing automated systems that predict and explain judicial outcomes.


%% file: methodology.tex
\section{Methodology}

\begin{figure}[t] 
    \centering 
    \includegraphics[width=\linewidth]{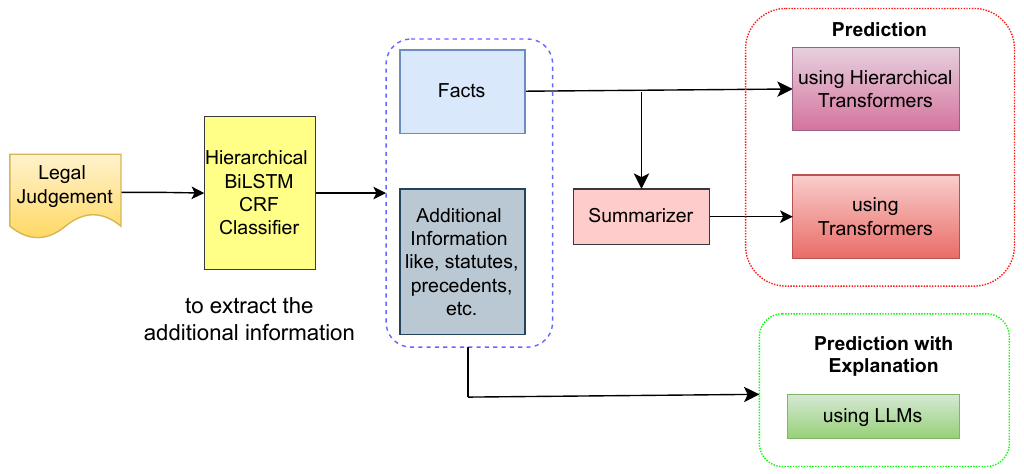}  
    \caption{Workflow for Legal Judgment Prediction with explanation.}
   \label{fig1}
\end{figure}

\subsection{Extraction of Facts and Additional Information from Judgments}
To extract relevant sentences from legal judgments, we employ a Hierarchical BiLSTM-CRF classifier, focusing on different rhetorical roles as identified by ~\cite{ghosh2019identification}. To create a realistic scenario for our model, we utilize the factual and additional contextual information such as statutes and precedents of the judgments as input for transformer models and LLMs.

\subsection{Transformer and Hierarchical Transformer Models}
The extracted facts undergo summarization using various techniques, including CaseSummarizer ~\cite{polsley2016casesummarizer}, BertSum ~\cite{liu2019fine}, SummaRuNNer ~\cite{nallapati2017summarunner}, and LetSum ~\cite{farzindar2004atefeh}, to ensure they fit within the input constraints of transformer models. Given that models like XLNET-large ~\cite{yang2019xlnet}, BERT ~\cite{devlin2018bert}, and InLegalBERT ~\cite{paul2022pre} can process a maximum input length of 512 tokens, we summarize the facts accordingly. Additionally, we utilize hierarchical transformer models that allow us to input the entire set of facts without the need for summarization. This approach facilitates the handling of comprehensive legal information during the prediction task, which is a binary classification problem.

\subsection{Prediction with Explanation using LLMs}
For the explanation task, we leverage LLMs such as Llama-2 (70b \& 13b) \cite{touvron2023llama} and GPT-3.5 Turbo \cite{brown2020language}, employing a prompting strategy. Given that the combined input and response length for these models is 4096 tokens, we segment the inputs into chunks of 2048 words. This segmentation allows us to generate judgment predictions, as one token corresponds to approximately three-quarters of a word, translating to about 750 words for 1000 tokens\footnote{\href{https://help.openai.com/en/articles/4936856-what-are-tokens-and-how-to-count-them}{what-are-tokens-and-how-to-count-them}}. We then aggregate the outputs from multiple chunks using a majority voting mechanism to determine the final judgment; in the event of a tie, the judgment is considered in favor of the appellant. For inputs shorter than 2048 words, we directly input the entire text into the LLM without requiring majority voting. We explore two prompting techniques:\\
\textbf{Normal Prompting:} The prompt states, ``You are asked to be a judge of a legal case and provide a judgment of the following legal judgment: \verb|<Legal judgment>|."\\
\textbf{Chain-of-Thought Prompting (CoT):} Following the chain-of-thought approach proposed by ~\cite{wei2022chain}, the prompt is modified to include, ``Think Step by Step."

We investigate six variations for each model input including sentences from:\\
\textbf{V1:} Only facts.\\
\textbf{V2:} V1 + statutes, and precedents.\\
\textbf{V3:} V2 + rulings by lower courts.\\
\textbf{V4:} V3 + arguments.\\
\textbf{V1+CoT:} Similar to V1, but incorporates the CoT prompt, ``Think Step by Step."\\
\textbf{V4+CoT:} Similar to V4, but includes the CoT.

Variations V1 and V2 simulate realistic scenarios where only essential elements, such as facts, statutes, and precedents, are provided to the LLM. These components mirror how judges typically approach cases by relying on the factual context and legal frameworks. V3 accounts for cases where a lower court has previously ruled on the matter, adding another layer of realism by simulating situations where an appeal is being heard. V4 enhances the prediction process by including arguments from legal counsel, simulating the complexity of real courtroom proceedings.

Prompting strategies engage both Task A (prediction) and Task B (explanation), thereby facilitating a comprehensive approach to judgment prediction and rationale generation.

%% file: results_and_analysis.tex
\section{Evaluation of Model Performance}
%
\subsection{Automatic Evaluation}
Table~\ref{tab:binary-classification-llm} summarizes the performance of judgment predictions made by different LLMs through prompting. The results demonstrate that relying solely on factual information leads to lower performance scores. However, incorporating additional legal case-specific information, such as statutes, precedents, rulings from lower courts, and arguments, significantly enhances the quality of predictions. Among the evaluated models, GPT-3.5 Turbo demonstrates the best overall performance.

\begin{table}[t]
\centering
\resizebox{\columnwidth}{!}{%
\begin{tabular}{lllllll}
\toprule
\textbf{Metric} & \textbf{V1} & \textbf{V2} & \textbf{V3} & \textbf{V4} & \textbf{V1+CoT} 
& \textbf{V4+CoT}    
\\ 
\midrule

\multicolumn{7}{c}{\textbf{Llama-2-13b}}
\\ 
\midrule
Precision  & 0.6443 & 0.6839 & 0.6941 & 0.6997 & 0.6821 & \textbf{0.7221} \\ 

Recall  & 0.6292 & 0.6246 & 0.6228 & 0.6416 & 0.6319 & \textbf{0.6824}   \\ 

F1-score & 0.6365 & 0.6528 & 0.6445 & 0.6693 & 0.6560 & \textbf{0.7016}     \\ 
\midrule
\multicolumn{7}{c}{\textbf{Llama-2-70b}}
\\ 
\midrule
Precision  & 0.7011 & 0.7344 & 0.7416 & \textbf{0.7518} & 0.7322 & 0.7416 \\ 

Recall  & 0.6644 & 0.6851 & 0.7147 & 0.6952 & 0.6817 & \textbf{0.7234}   \\ 

F1-score  & 0.6822 & 0.7088 & 0.7278 & 0.7223 & 0.7059 & \textbf{0.7323}   \\ 
\midrule
\multicolumn{7}{c}{\textbf{GPT-3.5 Turbo}} 
\\ 
\midrule
Precision  & 0.7016 & 0.7014 & 0.7411 & 0.7609 & 0.7261 & \textbf{0.7687} \\ 

Recall  & 0.6894 & 0.6914 & 0.6949 & \textbf{0.7155} & 0.6847 & 0.7132   \\ 

F1-score  & 0.6953 & 0.6962 & 0.7172 & 0.7374 & 0.7047 & \textbf{0.7398}   \\ 
\bottomrule
\end{tabular}
}
\caption{Performance Metrics for the Judgment Prediction Task on the ILDC-multi dataset using different LLMs across various input configurations (V1, V2, V3, V4, V1+CoT, V4+CoT), utilizing both normal prompting and CoT prompting. Bold values indicate the highest score for each metric and model.}
\label{tab:binary-classification-llm}
\end{table}

\begin{table}[t]
\centering
\resizebox{\columnwidth}{!}{%
\begin{tabular}{llllll}
\toprule
\textbf{Metric} & \textbf{HT} & \textbf{CS} & \textbf{SR} & \textbf{BS} & \textbf{LS}  
\\ 
\midrule

\multicolumn{6}{c}{\textbf{XLNET-large}}
\\ 
\midrule
Precision  & 0.6424 & 0.6313 & \textbf{0.6478} & 0.6227 & 0.5778 \\ 

Recall  & \textbf{0.6036} & 0.5713 & 0.5472 & 0.5683 & 0.5602 \\ 

F1-score & \textbf{0.6223} & 0.5998 & 0.5993 & 0.5942 & 0.5689   \\ 
\midrule
\multicolumn{6}{c}{\textbf{InlegalBERT}} 
\\ 
\midrule

Precision  & 0.6534 & 0.6415 & 0.6338 & \textbf{0.6604} & 0.6010 \\ 
Recall  & \textbf{0.6202} & 0.5673 & 0.5613 & 0.5885 & 0.5532    \\ 

F1-score & \textbf{0.6363} & 0.6022 & 0.5954 & 0.6223 & 0.5761     \\ 
\midrule
\multicolumn{6}{c}{\textbf{BERT}} 
\\ 
\midrule
Precision  & \textbf{0.6039} & 0.5557 & 0.5589 & 0.5592 & 0.5475 \\ 
Recall  & \textbf{0.5838} & 0.5540 & 0.5589 & 0.5589 & 0.5457   \\ 

F1-score  & \textbf{0.5936} & 0.5548 & 0.5589 & 0.5590 & 0.5466    \\ 
\bottomrule

\end{tabular}
}
\caption{Comparative Performance of Transformer Models on the Judgment Prediction Task on the ILDC-multi Dataset. These models are with fact summarization techniques such as CaseSummarizer (CS), SummaRuNNer (SR), BertSum (BS), and LetSum (LS), as well as Hierarchical Transformer (HT) models using the complete facts. Bold values indicate the highest score for each metric and model.}
\label{tab:comparative-metrics-transformer}
\end{table}

Table \ref{tab:comparative-metrics-transformer} provides further insights into the performance of various hierarchical transformer models and other transformer architectures. The results show that hierarchical transformer models outperform traditional summarization methods. Notably, models specifically pre-trained on Indian legal data, such as InlegalBERT, exhibit superior performance compared to those trained on generic datasets like BERT. The results indicate that LLMs have yet to reach the performance level of legal experts, who demonstrate a 94\% agreement rate, as noted by ~\cite{malik-etal-2021-ildc}.

\subsection{Expert Evaluation}
For the expert evaluation, we selected 25 explanations generated by the GPT-3.5 Turbo model, corresponding to different judgments, and enlisted three legal experts to assess these outputs. Each expert rated the explanations on a scale of 1 to 5 based on two criteria: (i) Clarity, the quality and coherence of the rationale behind the legal judgment, and (ii) Linking, the degree to which the explanation is logically connected to the final outcome of the judgment.

To ensure consistency and reliability in the evaluation, the experts were provided with clear guidelines. They were first instructed to familiarize themselves with both the legal judgments and the model-generated outputs to ensure informed assessments. For each explanation, they evaluated:

\textbf{Clarity:} This criterion focuses on how well the rationale is presented. A clear explanation should have a logical flow, use appropriate terminology, and be easily understood by both legal professionals and laypeople. The experts were asked to consider whether the explanation was coherent and if the reasoning behind the judgment was easy to follow.

\textbf{Linking:} This metric captures how well the explanation ties back to the final outcome. A strong linking score indicates that the rationale clearly leads to the conclusion of the judgment, without any gaps or inconsistencies. The experts were tasked with identifying whether the explanation logically and explicitly supports the final decision. The evaluators used the following rating scales:
\begin{itemize}
    \item \textbf{For Clarity:}\\
        \textbf{[1]:} Very Poor (Unclear rationale)\\
        \textbf{[2]:} Poor (Some clarity but weak rationale)\\
        \textbf{[3]:} Fair (Moderately clear rationale)\\
        \textbf{[4]:} Good (Clear rationale)\\
        \textbf{[5]:} Excellent (Very clear rationale)\\

    \item \textbf{For Linking:}\\
        \textbf{[1]:}Very Poor (Unclear and disconnected explanation)\\
        \textbf{[2]:}Poor (Weak linkage between explanation and judgment)\\
        \textbf{[3]:}Fair (Moderate linking, some gaps)\\
        \textbf{[4]:} Good (Clear linkage to the judgment)\\
        \textbf{[5]:}Excellent (Strong and coherent linking)\\
\end{itemize}

These ratings, calculated as the average scores for each criterion across the three experts, are presented in Table~\ref{tab:human_evaluation_gpt}. To ensure objectivity and ethical standards, the experts were instructed to maintain impartiality and avoid conflicts of interest throughout the evaluation process.

The results indicate that Variation 4 with chain-of-thought prompting (V4+CoT) achieved the highest scores for both clarity and linking, demonstrating its effectiveness in producing coherent and well-connected explanations. The average Fleiss' Kappa scores for Clarity and Linking were 0.64 and 0.70, respectively, indicating substantial agreement among the evaluators.

The combination of automatic and human evaluations offers a comprehensive assessment of the models' performance, revealing areas for improvement and confirming the efficacy of specific prompting techniques—such as chain-of-thought (CoT) in enhancing the quality of legal judgment prediction and explanation.

\begin{table}[t]
\centering
\resizebox{\columnwidth}{!}{%
\begin{tabular}{l rrrrrr}
\toprule
\textbf{Metric} & \textbf{V1} & \textbf{V2} & \textbf{V3} & \textbf{V4} & \textbf{V1+ CoT} & \textbf{V4+ CoT}
\\ 
\midrule
\bf Clarity  & 3.13 & 3.20 & 3.33 & 3.47 & 3.20 & \textbf{3.73} \\

\bf Linking  & 3.66 & 3.80 & 3.87 & 4.00 & 3.73 & \textbf{4.27}   \\
\bottomrule
\end{tabular}
}
\caption{Expert Evaluation Results for the Explanation Task Using GPT-3.5 Turbo. Bold values indicate the highest scores for each metric.}
\label{tab:human_evaluation_gpt}
\end{table}

%% file: conclusion_and_future_work.tex
\section{Conclusions}
In this study, we explored the effectiveness of various LLMs and transformer architectures in the task of judgment prediction and explanation using the ILDC-multi dataset. Our results demonstrate that incorporating additional case-specific information significantly enhances the prediction accuracy compared to using only factual information. The results also highlight the superiority of hierarchical transformer models over traditional summarization techniques, suggesting that a comprehensive approach to input data yields better predictive outcomes. Despite the promising results, our evaluations reveal that automated metrics still fall short of matching the performance levels of human legal experts, who demonstrate a high degree of agreement in judgment assessments. This gap underscores the need for further refinement of LLMs and transformer models to improve their interpretability and reliability in legal contexts.

%% file: limitations.tex
\section*{Limitations}

This study is focused solely on Supreme Court of India (SCI) judgments, which may limit the generalizability of the models to other courts or jurisdictions. Legal systems in different countries, or even lower courts within the same system, may have distinct structures, procedures, and nuances that are not captured in this study.

Additionally, the judgment prediction task is simplified as a binary classification problem. In real-world cases, particularly in multi-issue appeals, an appellant may win on some points and lose on others. This complexity is not fully addressed here, as our model reduces the outcome to a binary decision, which may overlook the nuances of cases with multiple heads of appeal.

While we incorporate facts, statutes, precedents, and arguments to simulate a realistic scenario, this approach still does not capture the full range of judicial reasoning. Judges often rely on implicit legal reasoning, judicial discretion, and a wider array of contextual factors that may not be explicitly mentioned in legal documents, limiting the comprehensiveness of our model’s predictions.

The large language models (LLMs) used in this study, such as GPT-3.5 Turbo and Llama-2, offer promising results, but their high computational requirements make them resource-intensive. This could restrict their practical application in many legal environments, especially in resource-constrained settings.

Furthermore, the human evaluation metrics—\textit{Clarity} and \textit{Linking}—are based on subjective assessments from legal experts. Although we provided detailed guidelines to standardize the evaluation process, differences in interpretation among experts can introduce variability into the results.

Future research will focus on addressing these limitations by exploring multi-label classification to account for more complex case outcomes, expanding the applicability of models to other legal domains and jurisdictions, and refining evaluation metrics to minimize subjectivity.

%% file: ethics.tex
\section*{Ethical Considerations}

In conducting this research, we adhered to ethical standards, particularly in the context of data usage and expert evaluation. The legal judgments used in our experiments were publicly available, and no private or sensitive data was accessed. For the human evaluation of judgment predictions and explanations, we engaged PhD scholars from the Rajiv Gandhi School of Intellectual Property Law as legal experts. Their participation was voluntary, and we provided monetary compensation for their time and expertise. This ensured that the evaluation process was both fair and conducted with proper acknowledgment of the experts' contributions.

%% file: appendix.tex
\newpage
\appendix

\section{Expert Evaluation}
\label{sec:Expert-Evaluation}
Table ~\ref{a1} shows scores provided by three legal experts for V1. Table ~\ref{a2} shows scores provided by three legal experts for V2.
Table ~\ref{a3} shows scores provided by three legal experts for V3.
Table ~\ref{a4} shows scores provided by three legal experts for V4. Table ~\ref{a5} shows scores provided by three legal experts for V1+CoT. Table ~\ref{a6} shows scores provided by three legal experts for V4+CoT.

Table ~\ref{a7} shows scores provided by three legal experts for V1. Table ~\ref{a8} shows scores provided by three legal experts for V2.
Table ~\ref{a9} shows scores provided by three legal experts for V3.
Table ~\ref{a10} shows scores provided by three legal experts for V4. Table ~\ref{a11} shows scores provided by three legal experts for V1+CoT. Table ~\ref{a12} shows scores provided by three legal experts for V4+CoT.

\begin{table*}[ht]
\centering
\resizebox{0.8\textwidth}{!}{%
\begin{tabular}{|c|c|c|c|}
\hline
\textbf{Document} & \textbf{Legal Expert 1} & \textbf{Legal Expert 2} & \textbf{Legal Expert 3} \\ \hline
Document 1  & 3  & 3  & 4  \\ \hline
Document 2  & 3  & 4  & 3  \\ \hline
Document 3  & 5  & 5  & 5  \\ \hline
Document 4  & 4  & 4  & 4  \\ \hline
Document 5  & 3  & 4  & 4  \\ \hline
Document 6  & 4  & 4  & 4  \\ \hline
Document 7  & 5  & 5  & 5  \\ \hline
Document 8  & 2  & 2  & 4  \\ \hline
Document 9  & 2  & 2  & 2  \\ \hline
Document 10 & 1  & 2  & 2  \\ \hline
Document 11 & 3  & 3  & 4  \\ \hline
Document 12 & 3  & 3  & 4  \\ \hline
Document 13 & 3  & 3  & 4  \\ \hline
Document 14 & 3  & 3  & 4  \\ \hline
Document 15 & 2  & 2  & 3  \\ \hline
Document 16 & 5  & 5  & 5  \\ \hline
Document 17 & 4  & 4  & 5  \\ \hline
Document 18 & 2  & 2  & 2  \\ \hline
Document 19 & 2  & 3  & 2  \\ \hline
Document 20 & 2  & 2  & 2  \\ \hline
Document 21 & 2  & 2  & 2  \\ \hline
Document 22 & 2  & 2  & 2  \\ \hline
Document 23 & 4  & 4  & 4  \\ \hline
Document 24 & 1  & 2  & 1  \\ \hline
Document 25 & 3  & 4  & 3  \\ \hline
\end{tabular}%
}

\caption{Clarity ratings from three legal experts in V1}
\label{a1}
\end{table*}

\newpage

\begin{table*}[ht]
\centering
\resizebox{0.8\textwidth}{!}{%
\begin{tabular}{|c|c|c|c|}
\hline
\textbf{Document} & \textbf{Legal Expert 1} & \textbf{Legal Expert 2} & \textbf{Legal Expert 3} \\ \hline
Document 1  & 4  & 4  & 4  \\ \hline
Document 2  & 3  & 4  & 4  \\ \hline
Document 3  & 3  & 4  & 3  \\ \hline
Document 4  & 5  & 5  & 5  \\ \hline
Document 5  & 2  & 2  & 2  \\ \hline
Document 6  & 2  & 3  & 3  \\ \hline
Document 7  & 4  & 4  & 4  \\ \hline
Document 8  & 2  & 2  & 2  \\ \hline
Document 9  & 3  & 3  & 3  \\ \hline
Document 10 & 3  & 3  & 3  \\ \hline
Document 11 & 5  & 5  & 5  \\ \hline
Document 12 & 3  & 3  & 3  \\ \hline
Document 13 & 2  & 2  & 2  \\ \hline
Document 14 & 3  & 3  & 4  \\ \hline
Document 15 & 4  & 4  & 4  \\ \hline
Document 16 & 2  & 2  & 2  \\ \hline
Document 17 & 2  & 2  & 3  \\ \hline
Document 18 & 5  & 5  & 5  \\ \hline
Document 19 & 3  & 3  & 4  \\ \hline
Document 20 & 4  & 4  & 4  \\ \hline
Document 21 & 1  & 2  & 2  \\ \hline
Document 22 & 2  & 2  & 3  \\ \hline
Document 23 & 3  & 3  & 4  \\ \hline
Document 24 & 2  & 2  & 3  \\ \hline
Document 25 & 5  & 5  & 5  \\ \hline
\end{tabular}%
}

\caption{Clarity ratings from three legal experts in V2}
\label{a2}
\end{table*}

\newpage

\begin{table*}[ht]
\centering
\resizebox{0.8\textwidth}{!}{%
\begin{tabular}{|c|c|c|c|}
\hline
\textbf{Document} & \textbf{Legal Expert 1} & \textbf{Legal Expert 2} & \textbf{Legal Expert 3} \\ \hline
Document 1  & 2  & 3  & 3  \\ \hline
Document 2  & 5  & 5  & 5  \\ \hline
Document 3  & 3  & 4  & 4  \\ \hline
Document 4  & 2  & 2  & 2  \\ \hline
Document 5  & 4  & 4  & 4  \\ \hline
Document 6  & 3  & 3  & 3  \\ \hline
Document 7  & 3  & 4  & 4  \\ \hline
Document 8  & 2  & 2  & 3  \\ \hline
Document 9  & 4  & 4  & 4  \\ \hline
Document 10 & 5  & 5  & 5  \\ \hline
Document 11 & 3  & 3  & 3  \\ \hline
Document 12 & 1  & 2  & 3  \\ \hline
Document 13 & 2  & 3  & 3  \\ \hline
Document 14 & 5  & 5  & 5  \\ \hline
Document 15 & 3  & 3  & 4  \\ \hline
Document 16 & 2  & 3  & 4  \\ \hline
Document 17 & 3  & 3  & 3  \\ \hline
Document 18 & 4  & 5  & 5  \\ \hline
Document 19 & 4  & 4  & 5  \\ \hline
Document 20 & 2  & 3  & 3  \\ \hline
Document 21 & 2  & 2  & 2  \\ \hline
Document 22 & 3  & 3  & 3  \\ \hline
Document 23 & 2  & 2  & 2  \\ \hline
Document 24 & 4  & 4  & 4  \\ \hline
Document 25 & 5  & 4  & 4  \\ \hline
\end{tabular}%
}

\caption{Clarity ratings from three legal experts in V3}
\label{a3}
\end{table*}

\newpage

\begin{table*}[ht]
\centering
\resizebox{0.8\textwidth}{!}{%
\begin{tabular}{|c|c|c|c|}
\hline
\textbf{Document} & \textbf{Legal Expert 1} & \textbf{Legal Expert 2} & \textbf{Legal Expert 3} \\ \hline
Document 1  & 5  & 5  & 5  \\ \hline
Document 2  & 3  & 3  & 4  \\ \hline
Document 3  & 3  & 3  & 4  \\ \hline
Document 4  & 5  & 5  & 5  \\ \hline
Document 5  & 2  & 2  & 3  \\ \hline
Document 6  & 2  & 2  & 3  \\ \hline
Document 7  & 5  & 5  & 5  \\ \hline
Document 8  & 3  & 3  & 3  \\ \hline
Document 9  & 3  & 3  & 3  \\ \hline
Document 10 & 3  & 3  & 3  \\ \hline
Document 11 & 2  & 3  & 3  \\ \hline
Document 12 & 4  & 4  & 4  \\ \hline
Document 13 & 3  & 4  & 3  \\ \hline
Document 14 & 3  & 4  & 4  \\ \hline
Document 15 & 5  & 5  & 5  \\ \hline
Document 16 & 4  & 4  & 5  \\ \hline
Document 17 & 2  & 3  & 3  \\ \hline
Document 18 & 4  & 4  & 4  \\ \hline
Document 19 & 3  & 3  & 3  \\ \hline
Document 20 & 2  & 3  & 3  \\ \hline
Document 21 & 2  & 2  & 3  \\ \hline
Document 22 & 3  & 3  & 4  \\ \hline
Document 23 & 4  & 4  & 4  \\ \hline
Document 24 & 4  & 4  & 4  \\ \hline
Document 25 & 2  & 2  & 2  \\ \hline
\end{tabular}%
}

\caption{Clarity ratings from three legal experts in V4}
\label{a4}
\end{table*}
\newpage

\begin{table*}[ht]
\centering
\resizebox{0.8\textwidth}{!}{%
\begin{tabular}{|c|c|c|c|}
\hline
\textbf{Document} & \textbf{Legal Expert 1} & \textbf{Legal Expert 2} & \textbf{Legal Expert 3} \\ \hline
Document 1  & 2  & 3  & 3  \\ \hline
Document 2  & 4  & 4  & 4  \\ \hline
Document 3  & 5  & 5  & 5  \\ \hline
Document 4  & 3  & 3  & 4  \\ \hline
Document 5  & 2  & 2  & 2  \\ \hline
Document 6  & 3  & 3  & 4  \\ \hline
Document 7  & 3  & 3  & 4  \\ \hline
Document 8  & 1  & 1  & 2  \\ \hline
Document 9  & 2  & 2  & 2  \\ \hline
Document 10 & 5  & 5  & 5  \\ \hline
Document 11 & 2  & 3  & 2  \\ \hline
Document 12 & 3  & 3  & 3  \\ \hline
Document 13 & 5  & 5  & 5  \\ \hline
Document 14 & 4  & 4  & 4  \\ \hline
Document 15 & 2  & 2  & 3  \\ \hline
Document 16 & 3  & 3  & 4  \\ \hline
Document 17 & 4  & 4  & 4  \\ \hline
Document 18 & 3  & 3  & 4  \\ \hline
Document 19 & 4  & 4  & 4  \\ \hline
Document 20 & 3  & 3  & 4  \\ \hline
Document 21 & 2  & 3  & 3  \\ \hline
Document 22 & 2  & 3  & 2  \\ \hline
Document 23 & 5  & 5  & 5  \\ \hline
Document 24 & 3  & 3  & 3  \\ \hline
Document 25 & 2  & 3  & 2  \\ \hline
\end{tabular}%
}

\caption{Clarity ratings from three legal experts for V1+CoT}
\label{a5}
\end{table*}

\newpage

\begin{table*}[ht]
\centering
\resizebox{0.8\textwidth}{!}{%
\begin{tabular}{|c|c|c|c|}
\hline
\textbf{Document} & \textbf{Legal Expert 1} & \textbf{Legal Expert 2} & \textbf{Legal Expert 3} \\ \hline
Document 1  & 5  & 5  & 5  \\ \hline
Document 2  & 2  & 3  & 2  \\ \hline
Document 3  & 4  & 4  & 5  \\ \hline
Document 4  & 4  & 4  & 5  \\ \hline
Document 5  & 3  & 4  & 3  \\ \hline
Document 6  & 4  & 5  & 4  \\ \hline
Document 7  & 4  & 4  & 4  \\ \hline
Document 8  & 2  & 2  & 2  \\ \hline
Document 9  & 3  & 4  & 3  \\ \hline
Document 10 & 4  & 4  & 5  \\ \hline
Document 11 & 5  & 5  & 5  \\ \hline
Document 12 & 3  & 3  & 4  \\ \hline
Document 13 & 4  & 4  & 5  \\ \hline
Document 14 & 2  & 3  & 3  \\ \hline
Document 15 & 4  & 4  & 4  \\ \hline
Document 16 & 2  & 2  & 3  \\ \hline
Document 17 & 4  & 4  & 4  \\ \hline
Document 18 & 3  & 3  & 4  \\ \hline
Document 19 & 5  & 5  & 5  \\ \hline
Document 20 & 4  & 4  & 4  \\ \hline
Document 21 & 2  & 2  & 3  \\ \hline
Document 22 & 3  & 3  & 4  \\ \hline
Document 23 & 5  & 5  & 5  \\ \hline
Document 24 & 2  & 2  & 3  \\ \hline
Document 25 & 4  & 4  & 4  \\ \hline
\end{tabular}%
}

\caption{Clarity ratings from three legal experts for V4+CoT}
\label{a6}
\end{table*}

\begin{table*}[ht]
\centering
\resizebox{0.8\textwidth}{!}{%
\begin{tabular}{|c|c|c|c|}
\hline
\textbf{Document} & \textbf{Legal Expert 1} & \textbf{Legal Expert 2} & \textbf{Legal Expert 3} \\ \hline
Document 1  & 3  & 4  & 4  \\ \hline
Document 2  & 3  & 4  & 3  \\ \hline
Document 3  & 5  & 5  & 5  \\ \hline
Document 4  & 3  & 4  & 3  \\ \hline
Document 5  & 5  & 5  & 4  \\ \hline
Document 6  & 3  & 4  & 3  \\ \hline
Document 7  & 4  & 5  & 5  \\ \hline
Document 8  & 3  & 4  & 4  \\ \hline
Document 9  & 3  & 4  & 1  \\ \hline
Document 10 & 3  & 3  & 2  \\ \hline
Document 11 & 3  & 3  & 4  \\ \hline
Document 12 & 3  & 3  & 4  \\ \hline
Document 13 & 4  & 4  & 4  \\ \hline
Document 14 & 5  & 5  & 4  \\ \hline
Document 15 & 3  & 3  & 3  \\ \hline
Document 16 & 4  & 5  & 4  \\ \hline
Document 17 & 4  & 4  & 5  \\ \hline
Document 18 & 5  & 5  & 2  \\ \hline
Document 19 & 4  & 5  & 2  \\ \hline
Document 20 & 3  & 3  & 2  \\ \hline
Document 21 & 4  & 5  & 2  \\ \hline
Document 22 & 4  & 5  & 1  \\ \hline
Document 23 & 4  & 4  & 4  \\ \hline
Document 24 & 4  & 4  & 1  \\ \hline
Document 25 & 4  & 4  & 3  \\ \hline
\end{tabular}%
}

\caption{Linking ratings from three legal experts for V1}
\label{a7}
\end{table*}

\newpage

\begin{table*}[ht]
\centering
\resizebox{0.8\textwidth}{!}{%
\begin{tabular}{|c|c|c|c|}
\hline
\textbf{Document} & \textbf{Legal Expert 1} & \textbf{Legal Expert 2} & \textbf{Legal Expert 3} \\ \hline
Document 1  & 3  & 4  & 4  \\ \hline
Document 2  & 3  & 4  & 4  \\ \hline
Document 3  & 5  & 5  & 5  \\ \hline
Document 4  & 3  & 4  & 4  \\ \hline
Document 5  & 5  & 5  & 5  \\ \hline
Document 6  & 3  & 4  & 4  \\ \hline
Document 7  & 4  & 4  & 4  \\ \hline
Document 8  & 3  & 4  & 3  \\ \hline
Document 9  & 3  & 4  & 4  \\ \hline
Document 10 & 3  & 4  & 4  \\ \hline
Document 11 & 3  & 4  & 4  \\ \hline
Document 12 & 3  & 3  & 3  \\ \hline
Document 13 & 4  & 4  & 4  \\ \hline
Document 14 & 5  & 5  & 5  \\ \hline
Document 15 & 3  & 4  & 4  \\ \hline
Document 16 & 4  & 4  & 3  \\ \hline
Document 17 & 4  & 4  & 4  \\ \hline
Document 18 & 5  & 5  & 5  \\ \hline
Document 19 & 4  & 5  & 5  \\ \hline
Document 20 & 3  & 3  & 3  \\ \hline
Document 21 & 4  & 4  & 3  \\ \hline
Document 22 & 4  & 4  & 3  \\ \hline
Document 23 & 3  & 3  & 2  \\ \hline
Document 24 & 3  & 3  & 2  \\ \hline
Document 25 & 3  & 3  & 3  \\ \hline
\end{tabular}%
}

\caption{Linking ratings from three legal experts for V2}
\label{a8}
\end{table*}

\newpage

\begin{table*}[ht]
\centering
\resizebox{0.8\textwidth}{!}{%
\begin{tabular}{|c|c|c|c|}
\hline
\textbf{Document} & \textbf{Legal Expert 1} & \textbf{Legal Expert 2} & \textbf{Legal Expert 3} \\ \hline
Document 1  & 4  & 5  & 4  \\ \hline
Document 2  & 4  & 4  & 3  \\ \hline
Document 3  & 5  & 5  & 4  \\ \hline
Document 4  & 4  & 4  & 4  \\ \hline
Document 5  & 5  & 5  & 5  \\ \hline
Document 6  & 4  & 4  & 4  \\ \hline
Document 7  & 4  & 4  & 3  \\ \hline
Document 8  & 4  & 4  & 4  \\ \hline
Document 9  & 4  & 4  & 4  \\ \hline
Document 10 & 4  & 4  & 3  \\ \hline
Document 11 & 4  & 4  & 3  \\ \hline
Document 12 & 3  & 3  & 3  \\ \hline
Document 13 & 4  & 4  & 4  \\ \hline
Document 14 & 5  & 5  & 5  \\ \hline
Document 15 & 4  & 4  & 3  \\ \hline
Document 16 & 4  & 4  & 4  \\ \hline
Document 17 & 4  & 4  & 4  \\ \hline
Document 18 & 5  & 5  & 3  \\ \hline
Document 19 & 5  & 5  & 4  \\ \hline
Document 20 & 3  & 3  & 3  \\ \hline
Document 21 & 4  & 4  & 4  \\ \hline
Document 22 & 4  & 2  & 3  \\ \hline
Document 23 & 3  & 2  & 2  \\ \hline
Document 24 & 3  & 2  & 2  \\ \hline
Document 25 & 3  & 2  & 2  \\ \hline
\end{tabular}%
}

\caption{Linking ratings from three legal experts for V3}
\label{a9}
\end{table*}

\begin{table*}[ht]
\centering
\resizebox{0.8\textwidth}{!}{%
\begin{tabular}{|c|c|c|c|}
\hline
\textbf{Document} & \textbf{Legal Expert 1} & \textbf{Legal Expert 2} & \textbf{Legal Expert 3} \\ \hline
Document 1  & 3  & 2  & 2  \\ \hline
Document 2  & 3  & 2  & 3  \\ \hline
Document 3  & 3  & 3  & 2  \\ \hline
Document 4  & 4  & 4  & 3  \\ \hline
Document 5  & 4  & 4  & 4  \\ \hline
Document 6  & 4  & 4  & 4  \\ \hline
Document 7  & 5  & 5  & 4  \\ \hline
Document 8  & 4  & 4  & 3  \\ \hline
Document 9  & 4  & 4  & 4  \\ \hline
Document 10 & 4  & 4  & 3  \\ \hline
Document 11 & 5  & 4  & 4  \\ \hline
Document 12 & 5  & 5  & 5  \\ \hline
Document 13 & 4  & 3  & 5  \\ \hline
Document 14 & 4  & 4  & 3  \\ \hline
Document 15 & 4  & 4  & 5  \\ \hline
Document 16 & 5  & 5  & 4  \\ \hline
Document 17 & 5  & 4  & 5  \\ \hline
Document 18 & 5  & 5  & 5  \\ \hline
Document 19 & 5  & 5  & 5  \\ \hline
Document 20 & 5  & 5  & 5  \\ \hline
Document 21 & 5  & 3  & 4  \\ \hline
Document 22 & 5  & 4  & 5  \\ \hline
Document 23 & 5  & 4  & 5  \\ \hline
Document 24 & 5  & 5  & 5  \\ \hline
Document 25 & 5  & 3  & 5  \\ \hline
\end{tabular}%
}

\caption{Linking ratings from three legal experts for V4}
\label{a10}
\end{table*}

\begin{table*}[ht]
\centering
\resizebox{0.8\textwidth}{!}{%
\begin{tabular}{|c|c|c|c|}
\hline
\textbf{Document} & \textbf{Legal Expert 1} & \textbf{Legal Expert 2} & \textbf{Legal Expert 3} \\ \hline
Document 1  & 2  & 3  & 3  \\ \hline
Document 2  & 4  & 4  & 4  \\ \hline
Document 3  & 5  & 5  & 5  \\ \hline
Document 4  & 3  & 3  & 4  \\ \hline
Document 5  & 2  & 2  & 2  \\ \hline
Document 6  & 3  & 3  & 4  \\ \hline
Document 7  & 3  & 1  & 4  \\ \hline
Document 8  & 1  & 1  & 2  \\ \hline
Document 9  & 2  & 2  & 5  \\ \hline
Document 10 & 5  & 5  & 2  \\ \hline
Document 11 & 2  & 3  & 2  \\ \hline
Document 12 & 3  & 3  & 3  \\ \hline
Document 13 & 5  & 5  & 5  \\ \hline
Document 14 & 4  & 4  & 4  \\ \hline
Document 15 & 2  & 2  & 4  \\ \hline
Document 16 & 3  & 3  & 5  \\ \hline
Document 17 & 4  & 4  & 4  \\ \hline
Document 18 & 3  & 3  & 4  \\ \hline
Document 19 & 4  & 3  & 4  \\ \hline
Document 20 & 3  & 3  & 4  \\ \hline
Document 21 & 2  & 3  & 3  \\ \hline
Document 22 & 2  & 3  & 2  \\ \hline
Document 23 & 5  & 5  & 5  \\ \hline
Document 24 & 3  & 3  & 2  \\ \hline
Document 25 & 2  & 3  & 5  \\ \hline
\end{tabular}%
}

\caption{Linking ratings from three legal experts for V1+CoT}
\label{a11}
\end{table*}

\begin{table*}[ht]
\centering
\resizebox{0.8\textwidth}{!}{%
\begin{tabular}{|c|c|c|c|}
\hline
\textbf{Document} & \textbf{Legal Expert 1} & \textbf{Legal Expert 2} & \textbf{Legal Expert 3} \\ \hline
Document 1  & 2  & 2  & 2  \\ \hline
Document 2  & 4  & 4  & 4  \\ \hline
Document 3  & 5  & 3  & 5  \\ \hline
Document 4  & 4  & 4  & 4  \\ \hline
Document 5  & 3  & 3  & 3  \\ \hline
Document 6  & 4  & 4  & 3  \\ \hline
Document 7  & 4  & 3  & 4  \\ \hline
Document 8  & 2  & 2  & 2  \\ \hline
Document 9  & 3  & 3  & 4  \\ \hline
Document 10 & 5  & 5  & 4  \\ \hline
Document 11 & 3  & 3  & 3  \\ \hline
Document 12 & 4  & 5  & 3  \\ \hline
Document 13 & 5  & 4  & 3  \\ \hline
Document 14 & 4  & 3  & 3  \\ \hline
Document 15 & 5  & 4  & 4  \\ \hline
Document 16 & 4  & 4  & 3  \\ \hline
Document 17 & 5  & 4  & 4  \\ \hline
Document 18 & 4  & 4  & 4  \\ \hline
Document 19 & 5  & 5  & 4  \\ \hline
Document 20 & 3  & 3  & 3  \\ \hline
Document 21 & 3  & 3  & 2  \\ \hline
Document 22 & 5  & 5  & 2  \\ \hline
Document 23 & 4  & 4  & 4  \\ \hline
Document 24 & 3  & 2  & 2  \\ \hline
Document 25 & 3  & 5  & 4  \\ \hline
\end{tabular}%
}

\caption{Linking ratings from three legal experts for V4+CoT}
\label{a12}
\end{table*}